\documentclass[runningheads]{llncs}

\usepackage{amsmath,amssymb,amsfonts,mathtools}
\usepackage{mathrsfs}
\usepackage{microtype}
\usepackage[hidelinks]{hyperref}
\usepackage{enumitem}
\usepackage{siunitx}
\usepackage{xcolor}
\usepackage{graphicx}
\usepackage{booktabs}
\usepackage{tabularx}
\usepackage{caption,subcaption}
\usepackage{comment}

\usepackage{tikz}
\usetikzlibrary{arrows.meta,positioning,shapes.geometric,fit,calc}

\usepackage{geometry}
\title{Perception Learning: A Formal Separation of Sensory Representation Learning from Decision Learning}
\titlerunning{Perception Learning}
\author{Suman Sanyal\inst{1}}
\authorrunning{S. Sanyal}
\institute{Big Data Analytics, Goa Institute of Management, Goa, India\\
\email{sanyal@gim.ac.in}}

\begin{document}
\maketitle

\begin{abstract}
We introduce Perception Learning (PeL), a paradigm that optimizes an agent's sensory interface $f_\phi:\mathcal{X}\to\mathcal{Z}$ using task-agnostic signals, decoupled from downstream decision learning $g_\theta:\mathcal{Z}\to\mathcal{Y}$. PeL directly targets label-free perceptual properties, such as stability to nuisances, informativeness without collapse, and controlled geometry, assessed via objective representation-invariant metrics. We formalize the separation of perception and decision, define perceptual properties independent of objectives or reparameterizations, and prove that PeL updates preserving sufficient invariants are orthogonal to Bayes task-risk gradients. Additionally, we provide a suite of task-agnostic evaluation metrics to certify perceptual quality. 
\end{abstract}

\section{Introduction}
Perception transforms raw sensory streams into internal codes that downstream decision modules can use. Contemporary ML often entangles perception and decision via end-to-end optimization. While powerful, this coupling can produce brittle, task-specific features and makes it hard to evaluate perception itself independently of any task head.

\begin{figure}[h!]
\centering
\scalebox{0.60}{
\begin{tikzpicture}[
  >=Latex,
  font=\small,
  node distance=8mm and 10mm,
  proc/.style={rectangle,rounded corners,draw,align=center,minimum width=3.3cm,minimum height=8mm},
  data/.style={trapezium,trapezium left angle=70,trapezium right angle=110,draw,align=center,minimum width=3.6cm,minimum height=8mm},
  note/.style={rectangle,draw,align=left,rounded corners,minimum width=3.3cm,minimum height=8mm},
  metric/.style={rectangle,draw,align=left,rounded corners,minimum width=4.2cm,minimum height=8mm},
  head/.style={rectangle,draw,align=center,rounded corners,minimum width=3.8cm,minimum height=8mm,thick},
  groupbox/.style={draw,rounded corners,dashed,inner sep=6pt},
  smallcap/.style={font=\footnotesize\itshape},
  xshift=-2.5cm
]
\node[data,xshift=-6cm] (datax) {Unlabeled data $X$\\(images, audio, text, state)};
\node[note, right=4cm of datax] (nuis) {Nuisances / group $G$:\\ rotations, lighting, time shift, \\ device, domain, etc.};
\draw[->] (datax) -- node[above,sloped,smallcap]{augment / sample views} (nuis);
\node[proc, below=of datax, xshift=-16mm] (view1) {view $x$};
\node[proc, below=of nuis,  xshift=-16mm] (view2) {view $x^{+}=T_\delta x$\\$\delta\sim\mathcal N(0,\sigma_{\text{aug}}^2)$};
\draw[->] (datax) -- (view1);
\draw[->] (nuis)  -- (view2);
\node[proc, below=23mm of $(view1)!0.5!(view2)$] (enc1) {Encoder $f_{\phi}$};
\node[proc, below=of enc1] (enc2) {Embedding $Z=f_{\phi}(x)$,\quad $Z^{+}=f_{\phi}(x^{+})$};
\node[note, right=12mm of enc1, yshift=7mm, text width=4.4cm] (invobj) {\textbf{Invariance / stability}\\ match $Z$ and $Z^{+}$ under $T_\delta$};
\node[note, right=12mm of enc2, yshift=-7mm, text width=4.4cm] (infoobj) {\textbf{Information / discriminability}\\ symmetric InfoNCE ($\tau$), avoid collapse};
\node[note, left=12mm of enc1, yshift=7mm, text width=4.4cm] (eqvobj) {\textbf{Equivariance}\\ optional $f(Tx)\!\approx\!\rho(T)f(x)$};
\node[note, left=12mm of enc2, yshift=-1mm, text width=4.4cm] (divobj) {\textbf{Diversity / anti-collapse}\\ variance/covariance floors};
\node[note, below=10mm of enc2, text width=5cm,xshift=-4.5cm] (leakobj) {\textbf{Nuisance leakage control}\\ minimize $I(Z;V)$ or adversarial heads};
\node[groupbox, fit=(enc1) (enc2) (invobj) (infoobj) (eqvobj) (divobj) (leakobj)] (pelbox) {};
\node[anchor=south west, smallcap] at ([xshift=-2mm,yshift=1.2mm]pelbox.north west) {Perception Learning (task-agnostic)};
\draw[->] (view1) -- (enc1.west);
\draw[->] (view2) -- (enc1.east);
\draw[-] (enc2.east) -- (invobj);
\draw[-] (enc2.east) -- (infoobj.west);
\draw[-] (enc2.west) -- (eqvobj);
\draw[-] (enc2.west) -- (divobj.east);
\draw[-] (enc2.west) -- (leakobj.north);
\draw[->] (enc1) -- (enc2);
\node[proc, below=18mm of pelbox] (zout) {Reusable representation $Z$};
\draw[->] (enc2) -- (zout);
\node[metric, below left=14mm and -1mm of zout, text width=5.4cm] (metrics) {Metrics on $Z$ (no labels):\\
$\bullet$ Invariance curve $D(\alpha)$, AUC\\
$\bullet$ Nuisance leakage (angle probe)\\
$\bullet$ Separability: Fisher, radial Fisher, MMD$^2$\\
$\bullet$ Label efficiency (linear probe; secondary)};
\draw[->] (zout.west) -- (metrics);
\node[head, right=20mm of zout, text width=4.2cm] (heads) {Decision Learning\\$g_{\theta}$ on $Z$\\(classification, control, QA, \dots)};
\draw[->] (zout) -- (heads);
\node[smallcap, text width=3.6cm, align=left,xshift=-2cm,yshift=5mm] (nograd) at ($(heads.north)!0.5!(heads.east)+(10mm,6mm)$) {no task gradients into $f_\phi$};
\draw[red!70, line width=0.6pt, -{Stealth[length=2mm]}] (heads.north) .. controls +(0,8mm) and +(12mm,8mm) .. (pelbox.north east);
\end{tikzpicture}
}
\caption{PeL trains $f_\phi$ on unlabeled views to produce a stable, informative, reusable code $Z$. Task heads $g_\theta$ are trained separately and do not backpropagate into $f_\phi$.}
\label{fig:pel-concept}
\end{figure}
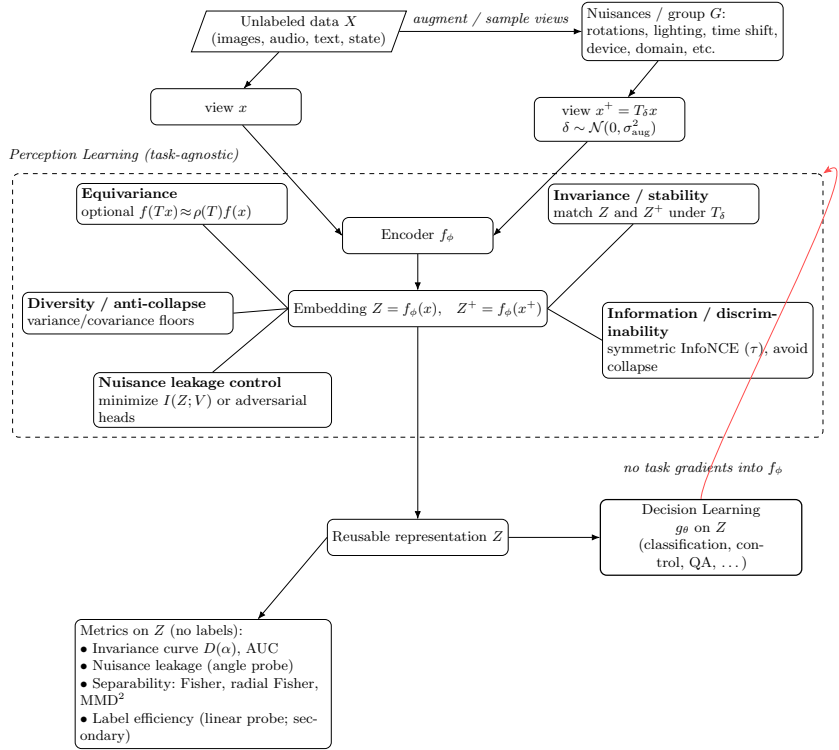

We introduce \textbf{Perception Learning} (PeL) which optimizes perception as a first-class objective using task-agnostic signals, while deferring decision learning to separate modules that consume the resulting code $Z$. Figure~\ref{fig:pel-concept} depicts the setup: PeL trains $f_\phi$ using invariance, information, and diversity signals on unlabeled data; decision heads $g_\theta$ are trained later and do not backpropagate into $f_\phi$. This separation yields reusable codes, clearer diagnostics, and safer invariance choices.

The remainder of this paper is organized as follows. Section~\ref{sec:related} surveys related work in cognitive science, self-supervised learning, equivariant networks, and modular systems, positioning PeL's unique contributions. Section~\ref{sec:pel-setup} formalizes the perception-decision split and defines PeL along with its minimal principles of separation, admissible supervision, and task-agnostic evaluation, as illustrated in Figures~\ref{fig:pel-in-agi} and~\ref{fig:pel-concept}. Section~\ref{sec:properties} introduces perceptual properties as representation-invariant functionals, presenting canonical examples such as invariance, information preservation, and geometric regularity, along with informal targets in Table~\ref{tab:pel-properties-min}. Section~\ref{sec:metrics} proposes a suite of task-agnostic metrics aligned with these properties, including invariance curves, leakage probes, and geometric diagnostics, summarized in Table~\ref{tab:pel-metrics}. Section~\ref{sec:theory} develops the theoretical foundation on standard Borel spaces, stating assumptions (A1)--(A5), proving a theorem on the orthogonality of PeL updates to Bayes task-risk gradients, and illustrating failure modes via counterexamples. We conclude in Section~\ref{sec:concl} with a summary, limitations, and directions for future work.

\begin{table}[t]
\centering
\caption{Key notations used in the paper.}
\label{tab:notations}
\small
\renewcommand{\arraystretch}{1.1}
\begin{tabular}{@{}ll@{}}
\toprule
\textbf{Notation} & \textbf{Meaning} \\
\midrule
$\mathcal{X}$ & Input (sensory) space \\
$\mathcal{Z}$ & Perceptual code space (Euclidean) \\
$\mathcal{Y}$ & Label/action space \\
$f_\phi: \mathcal{X} \to \mathcal{Z}$ & Sensory encoder (perception) \\
$g_\theta: \mathcal{Z} \to \Delta(\mathcal{Y})$ & Decision head (to simplex over $\mathcal{Y}$) \\
$Z = f_\phi(X)$ & Learned representation/code \\
$P_X$ & Input distribution \\
$G$ & Family of admissible transforms/nuisances \\
$T_\delta \in G$ & Sampled transform (e.g., augmentation) \\
$\pi: \mathcal{X} \to \mathcal{X}/G$ & Orbit map (to quotient space) \\
$T = \pi(X)$ & $G$-invariant sufficient statistic \\
$\mu_G$ & Probability measure on $G$ \\
$\Phi_{\mathsf{P}}(f_\phi; P_X, G)$ & Functional for property $\mathsf{P}$ (target $\mathcal{T}_{\mathsf{P}} \subseteq \mathbb{R}$) \\
$L_{\mathrm{perc}}(\phi)$ & PeL objective/loss \\
$\beta, \tau, \gamma, \varepsilon$ & Hyperparameters (e.g., invariance weight, temperature, variance floor, threshold) \\
$\mathcal{R}(\phi, \theta)$ & Task risk under proper loss $\ell$ \\
$F(\phi)$ & Bayes risk through $f_\phi$ \\
$\sigma(Z)$ & $\sigma$-algebra generated by $Z$ \\
$\mathcal{M}$ & Manifold of injective factorizations through $T$ \\
$\eta(x) = \mathbb{P}(Y \in \cdot \mid X = x)$ & Task posterior (G-invariant under A1) \\
$\mathrm{D}_v F(\phi_0)$ & Directional derivative of $F$ along $v$ at $\phi_0$ \\
\bottomrule
\end{tabular}
\end{table}

\section{Related Work and Positioning}\label{sec:related}
Classic work in cognitive science established that perceptual abilities improve through experience and practice, independent of explicit task rewards. Gibson’s monograph articulated developmental mechanisms for perceptual differentiation~\cite{Gibson1969}, and Goldstone’s survey synthesized enduring effects such as attention weighting, unitization, and differentiation~\cite{Goldstone1998}. These lines motivate treating perception as a trainable capability in its own right. Representation learning, on the other hand, aims to learn useful feature spaces without task labels~\cite{Bengio2013}. Contrastive and predictive SSL methods like CPC~\cite{Oord2018CPC}, SimCLR~\cite{Chen2020SimCLR}, MoCo~\cite{He2020MoCo}, BYOL~\cite{Grill2020BYOL}, and reconstruction/predictive approaches like MAE~\cite{He2022MAE}, DINOv2~\cite{Oquab2023DINOv2} deliver broad downstream gains across tasks. Unifying, modality-agnostic objectives include data2vec~\cite{Baevski2022data2vec}, while vision–language pretraining like CLIP~\cite{Radford2021CLIP} shows that weak supervision at scale can yield general, reusable perception. More recently, I-JEPA~\cite{Assran2023IJEPA} reframes pretraining as predicting target representations from context representations instead of pixel-level generation, explicitly pushing for semantic invariances. Across these threads, evaluation is still largely cast through downstream task accuracy, not perception. A complementary line imposes structure on features via group equivariance in G-CNNs~\cite{CohenWelling2016GCNN}, 3D/SE(3) symmetry-aware networks~\cite{Fuchs2020SE3}, and invariance-based generalization principles have been explored via Invariant Risk Minimization (IRM)~\cite{Arjovsky2019IRM}. Our program leverages such a structure but evaluates it at the perceptual interface, decoupled from any particular decision head.

In terms of applications, robotics and vision emphasized that perception improves when agents act to gather informative views (active perception)~\cite{Bajcsy1988Active}. While compatible with active data acquisition, our focus is on perception-centric training and certification that remain conceptually separate from policy optimization. World Models~\cite{HaSchmidhuber2018WorldModels}, DreamerV3~\cite{Hafner2023DreamerV3} foreground model-based prediction before control, and Perceiver/Perceiver-IO provide modality-agnostic perception backbones~\cite{Jaegle2021Perceiver,Jaegle2021PerceiverIO}. These support the view that a strong, reusable perceptual stack benefits many tasks and embodiments, aligning with our separation of perception learning from downstream decision learning.

\begin{figure}[h!]
\centering
\scalebox{0.6}{
\begin{tikzpicture}[
  >=Latex,
  font=\small,
  node distance=10mm and 12mm,
  spine/.style={rectangle,rounded corners,draw,align=center,minimum width=5cm,minimum height=9mm},
  box/.style={rectangle,rounded corners,draw,align=center,minimum width=4.6cm,minimum height=9mm},
  data/.style={trapezium, trapezium left angle=70, trapezium right angle=110, draw, align=center, minimum width=5cm, minimum height=9mm},
  store/.style={rectangle,draw,align=center,minimum width=4.6cm,minimum height=9mm,rounded corners=2pt,fill=black!3},
  groupbox/.style={draw,rounded corners,dashed,inner sep=6pt},
  anno/.style={font=\footnotesize\itshape},
  line/.style={-Latex}
]
\node[data] (env) {Environment $\mathcal{E}$\\(world, web, tools)};
\node[spine, below=of env] (sensors) {Sensors / Data streams\\(vision, audio, text, state)};
\draw[line] (env) -- (sensors);
\node[spine, below=of sensors] (pelenc) {Encoder $f_{\phi}$};
\node[spine, below=3mm of pelenc] (pelobj) {PeL objectives: invariance, contrastive info, diversity, leakage control};
\node[groupbox, fit=(pelenc) (pelobj)] (pelbox) {};
\node[anno, anchor=south east] at ([xshift=-7.0cm,yshift=0.0mm]pelbox.north east) {Perception Learning};
\draw[line] (sensors) -- node[right,anno,yshift=+0.2cm]{unlabeled multi-modal $X$} (pelenc);
\draw[-] (pelenc) -- (pelobj);
\node[spine, below=of pelbox] (zrep) {Reusable representation $Z=f_{\phi}(X)$};
\draw[line] (pelenc) -- (zrep);
\node[spine, below=of zrep] (plan) {Planning / Decision learning $g_{\theta}$\\(RL, control, classification, reasoning)};
\draw[line] (zrep) -- (plan);
\node[spine, below=of plan] (policy) {Policy $\pi$ / Actuators};
\draw[line] (plan) -- (policy);
\draw[line, looseness=1.2, shorten >=2pt, shorten <=2pt]
  (policy.west) to[out=180, in=180]
  node[left,anno,pos=0.55]{actions $a$}
  (env.west);
\node[box, right=22mm of zrep] (wm) {World model $p_{\psi}$\\(dynamics / prediction over $Z$)};
\draw[line] (zrep.east) -- (wm.west);
\draw[line] (wm.south) |- node[pos=0.25, right,anno]{imagined rollouts / targets} (plan.east);
\node[store, right=22mm of plan, yshift=-1.5cm] (retr) {Retrieval \& Knowledge\\(memory, RAG, tools)};
\draw[line] (zrep.east) -- (retr.north);
\draw[line] (retr.west) -| node[above,anno,xshift=+1.1cm,yshift=+0.05cm]{context} (plan.south east);
\node[store, left=22mm of plan] (heads) {Task heads / Adapters\\(downstream modules)};
\draw[line] (zrep.west) -- (heads.north east);
\draw[line] (heads.east) -- (plan.west);
\draw[red!70, dashed, -{Latex[length=2mm]}] (plan.north east) .. controls +(0:10mm) and +(0:10mm) .. node[above,anno]{no task gradients} (pelbox.east);
\end{tikzpicture}
}
\caption{PeL in a broader AGI stack. $Z$ supports world modeling, planning/decision learning, retrieval/knowledge, and task heads. Task/return gradients do not flow into $f_\phi$.}
\label{fig:pel-in-agi}
\end{figure}
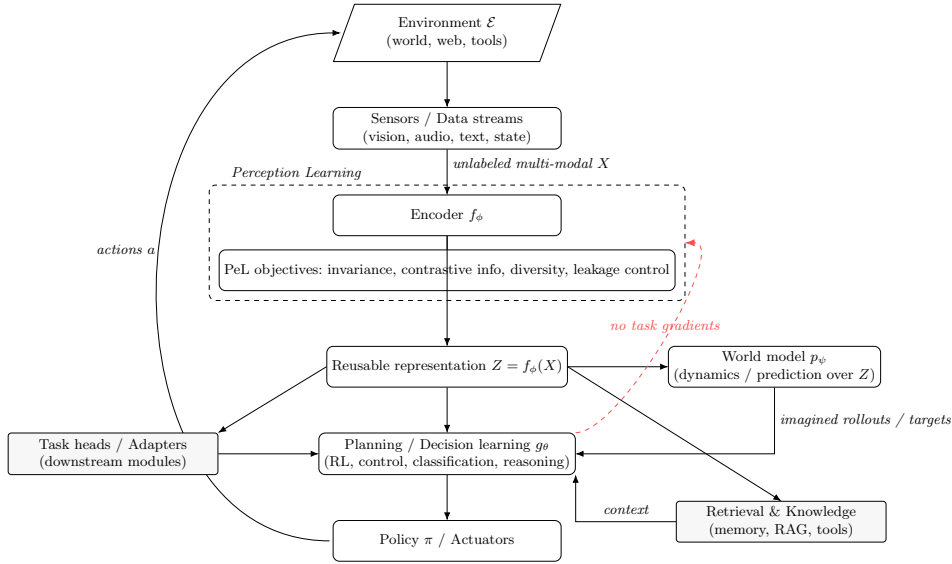

While Perception Learning (PeL) is adjacent to self-supervised learning, equivariance, world models, and active perception, but differs in three ways. 
\begin{enumerate}[label=(\roman*)]
\item PeL explicitly separates optimization of perception ($f_{\phi}\!:\!\mathcal{X}\!\to\!\mathcal{Z}$) from decision ($g_{\theta}\!:\!\mathcal{Z}\!\to\!\mathcal{Y}$),
\item PeL introduces task-agnostic, perception-first metrics that quantify improvement independently of downstream accuracy, and 
\item PeL provides training protocols that can be combined with, but are not subsumed by, decision learning. 
\end{enumerate}
In short, PeL operationalizes ``make perception first-class'' with concrete objectives and benchmarks that certify perception quality apart from policy or decoder performance. Refer to Fig.~\ref{fig:pel-in-agi} for schematics of PeL in an AGI-oriented stack.

\section{Perception vs Decision}\label{sec:pel-setup}
Let $x\!\in\!\mathcal{X}$ be raw input, $z\!\in\!\mathcal{Z}$ a perceptual code, and $y\!\in\!\mathcal{Y}$ a label/action. We posit
\begin{align}
z &= f_\phi(x) \quad \text{(perception)}, \\
\hat{y} &= g_\theta(z) \quad \text{(decision)}.
\end{align}
PeL optimizes $f_\phi$ with a label-free loss $L_{\mathrm{perc}}$; decision learning optimizes $g_\theta$ via a task loss $L_{\mathrm{task}}$ on frozen $f_\phi$.
\begin{definition}[Perception Learning]
\label{def:pel}
Perception learning is the problem of learning a sensory interface $f_\phi:\mathcal{X}\to\mathcal{Z}$ using task-agnostic signals such that $Z=f_\phi(X)$ acquires specified perceptual properties, without optimizing any task/decision loss.
\end{definition}
The minimal principles of PeL emphasize
\begin{enumerate}[label=(\roman*)]
\item separation, ensuring no task gradients flow to $\phi$, 
\item admissible supervision, restricted to augmentations, temporal proximity, predictive or reconstruction targets, and weak metadata, and 
\item evaluation, where success is measured by task-agnostic metrics on $Z$, with downstream accuracy considered secondary.
\end{enumerate}

\section{Perceptual Properties}\label{sec:properties}
The perceptual properties characterize the sensory interface independently of any downstream task. They can be used as training objectives, regularizers, or evaluation criteria. Typical choices target stability to nuisances, informativeness under capacity constraints, and geometric regularity of $Z=f_\phi(X)$.
\begin{definition}[Perceptual Property]
\label{def:perceptual-property}
Let $P_X$ be a distribution on $\mathcal{X}$, $G$ a family of admissible transforms on $\mathcal{X}$, and $f_\phi:\mathcal{X}\!\to\!\mathcal{Z}$ an encoder. A perceptual property $\mathsf{P}$ is a label-agnostic specification of the behavior of $f_\phi$ with respect to $(P_X,G)$ that is expressible as a measurable functional 
$\Phi_{\mathsf{P}}(f_\phi;P_X,G)\in\mathbb{R}$ with a target set $\mathcal{T}_{\mathsf{P}}\subseteq\mathbb{R}$. We say $f_\phi$ satisfies $\mathsf{P}$ when $\Phi_{\mathsf{P}}(f_\phi;P_X,G)\in\mathcal{T}_{\mathsf{P}}$. 
Crucially, $\Phi_{\mathsf{P}}$ depends only on $(X,G)$ and $f_\phi$, not on task labels, and is representation-invariant under injective reparameterizations $h:\mathcal{Z}\!\to\!\mathcal{Z}'$, i.e., if $f_\phi$ satisfies $\mathsf{P}$ then so does $h\!\circ\!f_\phi$.
\end{definition}

To formalize the goals of PeL, we define a set of perceptual properties that characterize desirable behaviors of the sensory interface $f_\phi: \mathcal{X} \to \mathcal{Z}$ with respect to the input distribution $P_X$ and a family of admissible transforms $G$. Each property is expressed as a measurable functional $\Phi_{\mathsf{P}}(f_\phi; P_X, G) \in \mathbb{R}$ with a target set $\mathcal{T}_{\mathsf{P}} \subseteq \mathbb{R}$, as per Definition 2, ensuring representation invariance under injective reparameterizations $h: \mathcal{Z} \to \mathcal{Z}'$. Below, we outline canonical classes of these properties with formal criteria, where $T_\delta \in G$ denotes a sampled transform, $V$ a known nuisance variable (when available), $\rho: G \to \mathrm{GL}(\mathcal{Z})$ a representation of $G$, and $d(\cdot,\cdot)$ a metric on $\mathcal{Z}$. For perceptual properties and informal targets we refer to Table~\ref{tab:pel-properties-min}.

The first property, \textbf{invariance}, ensures stability to nuisances, requiring that the representation $Z = f_\phi(x)$ remains consistent under transformations $T_\delta$. It is quantified as $\Phi_{\mathrm{inv}}(f_\phi) = \mathbb{E}_{x,\delta} \|f_\phi(x) - f_\phi(T_\delta x)\|_2^2 \le \varepsilon$, capturing the expected squared distance between embeddings of original and transformed inputs. \textbf{Equivariance}, in contrast, demands a structured response where transformations in $\mathcal{X}$ correspond to known transformations in $\mathcal{Z}$, defined as $\Phi_{\mathrm{eq}}(f_\phi) = \mathbb{E}_{x,\delta} \|f_\phi(T_\delta x) - \rho(\delta) f_\phi(x)\|_2^2 \le \varepsilon$, ensuring the encoder respects the group action $\rho$. 

\textbf{Information} preservation under capacity constraints is critical to avoid degenerate representations. This is formalized as $\Phi_{\mathrm{info}}(f_\phi) = -I(X; Z)$, subject to $\dim(Z) \le k$ or $\mathbb{E}\|f_\phi(x) - f_\phi(T_\delta x)\|^2 \le \varepsilon$, maximizing mutual information while respecting dimensionality or stability limits. Alternatively, a contrastive lower bound via InfoNCE is used: $\Phi_{\mathrm{info}}^{\mathrm{NCE}}(f_\phi) = \mathbb{E} \left[ -\log \frac{\exp(\mathrm{sim}(z, z^+)/\tau)}{\sum_{z^-} \exp(\mathrm{sim}(z, z^-)/\tau)} \right]$, where $z = f_\phi(x)$, $z^+ = f_\phi(T_\delta x)$, ensuring discriminability without collapse. \textbf{Nuisance independence}, or leakage control, minimizes the representation's sensitivity to nuisance variables, given by $\Phi_{\mathrm{leak}}(f_\phi) = I(Z; V)$ or a probe risk $\mathcal{R}_{\mathrm{probe}}(V \leftarrow Z) \le \varepsilon$, ensuring $Z$ discards irrelevant information.

\textbf{Geometric regularity} promotes smooth and well-conditioned embeddings, captured through multiple functionals: $\Phi_{\mathrm{lip}}(f_\phi) = \mathbb{E} \|\nabla_x f_\phi(x)\|_F^2$ for smoothness via Jacobian energy, $\Phi_{\mathrm{cov}}(f_\phi) = \sum_{i \neq j} (\mathrm{Cov}(Z_i, Z_j))^2$ for decorrelation, and $\Phi_{\mathrm{var}}(f_\phi) = \sum_d \max(0, \gamma - \mathrm{Var}(Z_d))$ to maintain variance and prevent collapse, each targeting small values. When generative factors $\mathcal{U}$ are known, \textbf{factor disentanglement} is defined as $\Phi_{\mathrm{dis}}(f_\phi) = \sum_{u \in \mathcal{U}} \min_d (1 - \mathrm{NMI}(Z_d, u))$ or total-correlation penalties on $Z$, encouraging alignment of representation dimensions with distinct factors. Finally, \textbf{sufficient} invariants ensure $Z$ captures all task-relevant information invariant to $G$, with a soft criterion $\Phi_{\mathrm{suff}}(f_\phi) = I(X; Z \mid \pi(X)) \le \varepsilon$, where $\pi: \mathcal{X} \to \mathcal{X}/G$ is an orbit map, meaning $Z$ carries minimal information beyond the invariant statistic.

These properties are evaluated using task-agnostic metrics tailored to each functional. Invariance is assessed via curves $D(\alpha) = \mathbb{E}\|f_\phi(x) - f_\phi(\tau_\alpha x)\|_2^2$ and their AUC, leakage via adversarial or linear probes, separability under information preservation via Fisher ratios or MMD$^2$, geometric properties via Jacobian or covariance diagnostics, and sufficiency through conditional mutual information surrogates. This comprehensive framework ensures that PeL produces representations that are stable, informative, and geometrically well-behaved, independently of downstream task objectives.

\begin{table}[t]
  \centering
  \caption{Perceptual properties and informal targets. “Code” is the learned representation \(Z=f_{\phi}(X)\).}
  \label{tab:pel-properties-min}
  \renewcommand{\arraystretch}{1.12}
  \scalebox{0.8}{
  \begin{tabular}{@{}p{6cm}p{9cm}@{}}
    \toprule
    \textbf{Perceptual Property} & \textbf{Informal target} \\
    \midrule
    \textbf{Stability / Invariance} &
    Same scene \(\Rightarrow\) similar code under admissible transforms \(T_\delta\). \\
    \addlinespace[0.6ex]
    \textbf{Equivariance} &
    Predictable change of the code under \(T\): applying \(T\) corresponds to a known transform \(\rho(T)\) in representation space. \\
    \addlinespace[0.6ex]
    \textbf{Information / Non-collapse} &
    Preserve information; avoid degenerate (constant) codes so distinct inputs remain distinguishable. \\
    \addlinespace[0.6ex]
    \textbf{Nuisance independence (leakage)} &
    Make the code insensitive to known nuisance variables (e.g., rotation angle, sensor ID). \\
    \addlinespace[0.6ex]
    \textbf{Geometric regularity} &
    Promote smooth, well-conditioned embeddings with controlled variance, decorrelation, and Jacobian energy. \\
    \addlinespace[0.6ex]
    \textbf{Factor disentanglement} &
    Align representation dimensions with distinct generative factors when known, minimizing total correlation. \\
    \addlinespace[0.6ex]
    \textbf{Sufficiency / Orbit statistic} &
    Retain invariant content (constant along \(G\)-orbits) and discard variation along nuisance directions. \\
    \bottomrule
  \end{tabular}
  }
\end{table}
\begin{remark}
While the canonical functionals for invariance and equivariance are defined using Euclidean $\ell_2$-norms, which are not strictly preserved under arbitrary injective reparameterizations $h: \mathcal{Z} \to \mathcal{Z}'$ (e.g., nonlinear scalings distort distances), the underlying property satisfaction remains equivalent: if $f_\phi$ meets the criterion, so does $h \circ f_\phi$ when the functional is adapted via a canonical metric (e.g., cosine similarity or kernel-induced distances). Properties based on information measures (e.g., $I(X; Z)$, $I(Z; V)$) are fully invariant due to $\sigma$-algebra preservation. For geometric terms like covariance penalties, affine $h$ suffice; in general, this mild assumption aligns with Euclidean representations in SSL literature, ensuring the framework's objectives are robust to downstream linear probes or heads.
\end{remark}

\section{Task-Agnostic Metrics for Perceptual Improvement}
\label{sec:metrics}

To evaluate the quality of the perceptual representation $Z = f_\phi(X)$ independently of downstream decision heads $g_\theta$, we propose a suite of task-agnostic metrics that directly assess the perceptual properties defined in Section 4. These metrics focus on stability, informativeness, nuisance independence, and geometric regularity, ensuring that improvements in perception are measured without conflating with task-specific performance.

\textbf{Nuisance independence} is quantified by estimating the mutual information $I(Z; V)$ between the representation $Z$ and known nuisance variables $V$ (e.g., rotation angles or sensor IDs). In practice, this can be computed using (i) adversarial probing, where a strong classifier attempts to predict $V$ from $Z$, with lower AUC indicating better independence, or (ii) nonparametric mutual information estimators. We report the normalized quantity $\widehat{I}(Z; V)/H(V)$ to contextualize leakage relative to the nuisance entropy, with lower values indicating better suppression of irrelevant information.

\textbf{Invariance} and \textbf{equivariance} are assessed through curves that measure the stability of $Z$ under a family of transformations $\tau_\alpha \in G$ (e.g., rotations parameterized by angle $\alpha$). We compute $D(\alpha) = \mathbb{E}[\|f_\phi(x) - f_\phi(\tau_\alpha(x))\|_2^2]$, summarizing invariance via the area-under-curve (AUC), where lower values indicate stronger stability to nuisances. For equivariance, similar curves can be adapted to measure alignment with the group action $\rho(\alpha)$, ensuring the representation responds predictably to transformations.

\textbf{Perceptual faithfulness} evaluates how well $Z$ captures the essential structure of the input $X$. If a decoder $p_\omega(X \mid Z)$ is available, we measure reconstruction quality via the expected negative log-likelihood $\mathbb{E}[-\log p_\omega(X \mid Z)]$. For specific modalities, metrics like PSNR or SSIM (for images) or log-spectral distance (for audio) provide practical proxies, reflecting the information preservation property without requiring task labels.

\textbf{Geometric regularity} is assessed through three metrics. First, local smoothness is measured as $S = \mathbb{E}[\|\nabla_x f_\phi(x)\|_F^2]$, capturing the Lipschitz continuity of the encoder to ensure stable responses to input perturbations. Second, when generative factors are known (e.g., in synthetic datasets), disentanglement scores, such as normalized mutual information (NMI) between $Z$ dimensions and factors, quantify axis-alignment. Third, the Fisher information trace $\mathrm{tr} \mathcal{I}_Z$ under small nuisance perturbations evaluates the representation’s sensitivity, with lower traces indicating robustness.

Finally, to gauge transfer readiness without conflating with decision learning, we employ a fixed-capacity linear probe on $Z$, trained with early stopping on a small validation split. This measures \textbf{data-efficiency} (accuracy vs. number of labels) but is treated as a secondary metric to avoid task bias. These metrics collectively ensure that $Z$ aligns with the perceptual properties of invariance, information preservation, nuisance independence, and geometric regularity, enabling robust evaluation of perception independent of downstream tasks.

\begin{table}[t]
  \centering
  \caption{Task-agnostic metrics for perceptual properties and their alignment.}
  \label{tab:pel-metrics}
  \renewcommand{\arraystretch}{1.12}
  \scalebox{0.8}{
  \begin{tabular}{p{8cm}p{6cm}}
    \toprule
    \textbf{Metric} & \textbf{Perceptual Property Targeted} \\
    \midrule
    Nuisance independence ($\widehat{I}(Z; V)/H(V)$ or probe AUC) &
    Nuisance independence (leakage control) \\
    \addlinespace[0.6ex]
    Invariance/equivariance curves ($D(\alpha)$, AUC) &
    Stability/invariance, Equivariance \\
    \addlinespace[0.6ex]
    Perceptual faithfulness ($\mathbb{E}[-\log p_\omega(X \mid Z)]$, PSNR/SSIM) &
    Information preservation \\
    \addlinespace[0.6ex]
    Local smoothness ($S = \mathbb{E}[\|\nabla_x f_\phi(x)\|_F^2]$) &
    Geometric regularity (smoothness) \\
    \addlinespace[0.6ex]
    Disentanglement scores (NMI for known factors) &
    Factor disentanglement \\
    \addlinespace[0.6ex]
    Fisher information trace ($\mathrm{tr} \mathcal{I}_Z$) &
    Geometric regularity (robustness to perturbations) \\
    \addlinespace[0.6ex]
    Linear probe data-efficiency (secondary) &
    Information preservation, Transfer readiness \\
    \bottomrule
  \end{tabular}
  }
\end{table}

\section{Separating Perception and Decision}\label{sec:theory}
Let the population task risk be
\begin{equation}
\label{eq:risk}
\mathcal{R}(\phi,\theta)=\mathbb{E}\big[\ell(g_\theta(f_\phi(X)),Y)\big].
\end{equation}
We say that $f_\phi$ is task-sufficient for $Y$ if there exists $h$ such that $\mathbb{P}(Y|X)=\mathbb{P}(Y|Z)$ with $Z=f_\phi(X)$ (i.e., $Z$ is sufficient for $Y$). Intuitively, if $f_\phi$ captures exactly the task-relevant (e.g., $G$-invariant) information, then further perception-only improvements that preserve this information should not degrade Bayes-optimal decision risk.
We now formalize the invariance setting and sufficiency conditions used in the full proof. We work on standard Borel spaces. Let $(\mathcal{X},\mathscr{X})$ be the input (sensory) space and $(\mathcal{Y},\mathscr{Y})$ the label/action space. A (measurable) group action $G \curvearrowright \mathcal{X}$ is written $(g,x)\mapsto g\!\cdot\!x$. Now consider the following assumptions. 
\begin{description}[leftmargin=1.1em]
\item[(A1) Group invariance of the target.]
There exists a (measurable) conditional distribution $\eta(x)\!=\!\mathbb{P}(Y\!\in\!\cdot\mid X\!=\!x)$ such that
$\eta(g\!\cdot\!x)=\eta(x)$ for all $g\in G$ and a.e.\ $x\in\mathcal{X}$. Equivalently, $Y$ depends on $X$ only through its $G$-orbit.
\item[(A2) Invariant (orbit) statistic and sufficiency.]
Let $\pi:\mathcal{X}\!\to\!\mathcal{T}$ be a measurable orbit map constant on $G$-orbits (the canonical projection to the quotient space $\mathcal{X}/G$). Assume the Markov condition
$Y \perp\!\!\!\perp X \mid T(X)$ with $T\!=\!\pi$. Thus $T$ is a $G$-invariant sufficient statistic
$\mathbb{P}(Y\!\in\!\cdot\mid X)=\mathbb{P}(Y\!\in\!\cdot\mid T(X))$.
\item[(A3) Perception and decision models.]
A differentiable representation $f_\phi:\mathcal{X}\!\to\!\mathcal{Z}$ and a decision $g_\theta:\mathcal{Z}\!\to\!\Delta(\mathcal{Y})$ (predictive distribution), trained with a strictly proper loss $\ell$ (log-loss, Brier, etc.), so the Bayes act for a given $Z$ is $g_\theta^\star(z)=\mathbb{P}(Y\!\in\!\cdot\mid Z\!=\!z)$. Assume $\mathcal{Z}$ is Euclidean, $f_\phi, g_\theta$ locally Lipschitz, with finite expectations under $\ell$, ensuring gradient interchanges via dominated convergence.
\item[(A4) Perception-invariance regularizer.]
Define
\[
L_{\mathrm{inv}}(\phi)
=\mathbb{E}\Big[\big\|f_\phi(X)-f_\phi(g\!\cdot\!X)\big\|_2^2\Big],
\quad g\sim \mu_G,
\]
for some probability measure $\mu_G$ on $G$. Its minima are (representations $f_\phi$ that are) $G$-invariant: $f_\phi(g\!\cdot\!x)=f_\phi(x)$.
\item[(A5) Factor-through-$T$ condition.]
We say $f_\phi$ factors through $T$ if $f_\phi=h\circ T$ for some measurable $h:\mathcal{T}\!\to\!\mathcal{Z}$.
We write $\sigma(Z)$ for the $\sigma$-algebra generated by a random variable $Z$.
\end{description}
\paragraph{Notation.}
Task risk $R(\phi,\theta)=\mathbb{E}\, \ell\!\big(g_\theta(f_\phi(X)),Y\big)$. The Bayes (task) risk through $f_\phi$ is $F(\phi):=\inf_\theta R(\phi,\theta)$. By strict propriety of $\ell$, $F(\phi)$ is achieved by $g^\star_\theta(z)=\mathbb{P}(Y\!\in\!\cdot\mid Z\!=\!z)$ and can be written as
$F(\phi)=\mathbb{E}\,L_\ell\!\big(\mathbb{P}(Y\!\in\!\cdot\mid Z)\big)$ for a convex functional $L_\ell$ determined by $\ell$.

\subsection{Orthogonality Under Group Invariance}
 The next result establishes that, under task-true group invariance, PeL updates that enhance stability without altering the information content of the invariant sufficient statistic are orthogonal to the Bayes task-risk gradient, justifying the separation of perception and decision learning.
\begin{theorem}[Orthogonality of PeL updates to Bayes-risk gradient]
\label{thm:orth}
Assume \textbf{(A1)--(A5)} and let $\phi_0$ satisfy $f_{\phi_0}=h_0\circ T$ with $h_0$ injective on $\mathrm{range}(T)$
(so $\sigma(f_{\phi_0}(X))=\sigma(T(X))$).
Let $v$ be any tangent direction at $\phi_0$ such that, for all sufficiently small $t$,
$f_{\phi_0+t v}=h_t\circ T$ for some measurable $h_t$ that remains injective on $\mathrm{range}(T)$.
Assume $\mathcal{M}$ is locally smooth, with $h_t$ continuous in $t$. Then:
\[
\mathrm{D}_v F(\phi_0)=0.
\]
In particular, if $\nabla_\phi L_{\mathrm{inv}}(\phi_0)$ is such a direction (i.e., it only improves $G$-invariance without altering $\sigma(T(X))$), then
\[
\nabla_\phi F(\phi_0)\,\cdot\,\nabla_\phi L_{\mathrm{inv}}(\phi_0)=0.
\]
\end{theorem}
\begin{proof}
By strict propriety of $\ell$ and the envelope theorem,
$F(\phi)=\mathbb{E}\,L_\ell\!\big(\mathbb{P}(Y\!\in\!\cdot\mid Z_\phi)\big)$ with $Z_\phi=f_\phi(X)$,
and the derivative $\mathrm{D}_v F(\phi_0)$ depends only on how the posterior
$\eta_\phi(z):=\mathbb{P}(Y\!\in\!\cdot\mid Z_\phi=z)$ varies with $\phi$ along $v$.
Under \textbf{(A2)} we have the conditional independence $Y\perp\!\!\!\perp X \mid T(X)$, hence
\[
\mathbb{P}(Y\!\in\!\cdot\mid Z_\phi)
= \mathbb{E}\!\left[\,\mathbb{P}(Y\!\in\!\cdot\mid T)\ \middle|\ Z_\phi\,\right]
= \mathbb{E}\!\left[\,\eta_T \ \middle|\ Z_\phi\,\right],
\]
where $\eta_T:=\mathbb{P}(Y\!\in\!\cdot\mid T)$.
If $f_\phi=h\circ T$ with $h$ injective on $\mathrm{range}(T)$, then $\sigma(Z_\phi)=\sigma(T)$
and thus
\[
\mathbb{P}(Y\!\in\!\cdot\mid Z_\phi)
= \mathbb{E}\!\left[\,\eta_T \ \middle|\ \sigma(Z_\phi)\,\right]
= \mathbb{E}\!\left[\,\eta_T \ \middle|\ \sigma(T)\,\right]
= \eta_T
\]
almost surely. Therefore $F(\phi)=\mathbb{E}\,L_\ell(\eta_T)$ is constant over the set of $\phi$ such that
$f_\phi$ factors through $T$ with an injective $h$ on $\mathrm{range}(T)$.
By the hypothesis on the tangent direction $v$, for all sufficiently small $t$, $f_{\phi_0+t v}=h_t\circ T$ with $h_t$ injective on $\mathrm{range}(T)$,
so the posterior $\mathbb{P}(Y\!\in\!\cdot\mid Z_{\phi_0+t v})$ equals $\eta_T$ for all small $t$, and hence $F(\phi_0+t v)$ is constant in $t$.
Thus $\mathrm{D}_v F(\phi_0)=0$.
For the last statement, if $\nabla_\phi L_{\mathrm{inv}}(\phi_0)$ is tangent to the same manifold
(i.e., it changes $h$ but keeps $\sigma(Z_{\phi})=\sigma(T)$), then by definition of directional derivative
\[
\nabla_\phi F(\phi_0)\cdot \nabla_\phi L_{\mathrm{inv}}(\phi_0) = \mathrm{D}_{-\nabla_\phi L_{\mathrm{inv}}} F(\phi_0)=0,
\]
establishing orthogonality.
\end{proof}
\begin{remark}
If the task is truly $G$-invariant (\textbf{A1}) and the representation already captures exactly the orbit information $T$ (no more, no less), then any further deformation of $f_\phi$ that keeps it a one-to-one reparametrization of $T$ leaves the Bayes posterior, and hence Bayes risk, unchanged. Perception-learning steps that only strengthen invariance (reduce within-orbit variance) without identifying distinct orbits collapse nothing task-relevant; they are therefore ``orthogonal'' to task-risk improvement.
\end{remark}
\begin{corollary}[Two-stage optimality under exact invariance]
\label{cor:two-stage}
Under \textbf{(A1)--(A5)}, if $f_{\phi^\star}=h^\star\circ T$ with $h^\star$ injective on $\mathrm{range}(T)$, then
\[
\inf_\theta R(\phi^\star,\theta)=\inf_{\theta} \mathbb{E}\,\ell\big(g_\theta(Z),Y\big) \quad \text{with}\ Z=T(X)
\]
achieves Bayes risk. Hence optimizing $f_\phi$ to (any) injective invariant parameterization of $T$ can be done prior to decision learning without sacrificing asymptotic task optimality.
\end{corollary}
\begin{proof}
Immediate from the proof of Theorem~\ref{thm:orth}. If $\sigma(Z)=\sigma(T)$ then
$\mathbb{P}(Y\!\in\!\cdot\mid Z)=\eta_T$ and the Bayes act attains Bayes risk.
\end{proof}

\subsection{Clarifying the Role of Assumptions}
Assumptions (A1)–(A2) establish the statistical backbone. (A1) states that the true posterior is constant on \(G\)-orbits; (A2) upgrades this to a sufficiency statement, asserting the existence of an invariant statistic \(T\) with \(Y \perp\!\!\!\perp X \mid T(X)\). Consequently, whenever \(\sigma(Z)=\sigma(T)\), conditioning on the representation \(Z=f_\phi(X)\) is equivalent to conditioning on \(T\), which lets the Bayes risk be expressed entirely through the \(G\)-invariant information.

Injectivity on \(\mathrm{range}(T)\) appears to prevent coarsening, i.e., if \(f_\phi=h\circ T\) with \(h\) not injective on \(\mathrm{range}(T)\), then \(Z=h(T)\) merges distinct orbits and may increase Bayes risk. We only require injectivity on the support of \(T\), which is weaker than global invertibility. Regarding the PeL update, when \(f_{\phi_0}\) is already near-invariant (small \(L_{\mathrm{inv}}\)), first-order updates that minimize \(L_{\mathrm{inv}}\) primarily reduce within-orbit variance and can be viewed as moving within (or tangent to) the manifold of \(T\)-preserving reparameterizations; as long as the update does not merge distinct orbits i.e., \(\sigma(Z)=\sigma(T)\) is preserved, it lies in the tangent cone used in Theorem~\ref{thm:orth}. This is precisely the PeL regime, strengthening invariance without losing orbit identity. Assumption (A3) provides a clean, differentiable Bayes-risk functional of the posterior. With a strictly proper loss and mild regularity,
\[
F(\phi)\;=\;\inf_\theta \mathcal{R}(\phi,\theta)
\;=\;\mathbb{E}\,L_\ell\!\big(\,\mathbb{P}(Y\!\in\!\cdot\,\mid\,Z_\phi)\,\big),
\]
and directional derivatives \(\mathrm{D}_v F(\phi)\) are well defined and compatible with envelope arguments. Assumption (A5) identifies the flat manifold of the Bayes risk
\[
\mathcal{M}\;=\;\bigl\{\phi:\; f_\phi = h\!\circ\! T,\ \text{$h$ injective on range}(T)\bigr\},
\]
i.e., all injective reparameterizations of the sufficient invariant \(T\). On \(\mathcal{M}\) we have \(\sigma(Z_\phi)=\sigma(T(X))\), hence \(F(\phi)\) is constant and \(\mathrm{D}_v F(\phi)=0\) for tangent directions \(v\).

Assumption (A4) supplies a concrete perception objective \(L_{\mathrm{inv}}\) whose gradient, near \(\mathcal{M}\), stays tangent to \(\mathcal{M}\). Therefore, at \(\phi_0\in\mathcal{M}\),
\[
\nabla_{\phi}F(\phi_0)\cdot\nabla_{\phi}L_{\mathrm{inv}}(\phi_0)=0,
\]
establishing the orthogonality of PeL updates to the Bayes task-risk gradient.

Taken together, (A1)–(A5) justify training perception to be (correctly) \(G\)-invariant without harming Bayes-optimal decision performance, and they delineate the failure modes when invariances are mis-specified or when updates merge distinct orbits.

\subsection{Counterexample: Over-Invariance Can Hurt}
We now show that enforcing too much invariance can strictly increase Bayes risk when (A1) fails (the label is not invariant) or when the update destroys orbit identity ($h$ not injective on $\mathrm{range}(T)$). Let $X=(U,V)$ with $U,V\in\{0,1\}$ i.i.d.\ Bernoulli($1/2$). Define the label $Y:=V$ (so $V$ is causal for $Y$ and not a nuisance). Let $G=\{\mathrm{id},\,\tau\}$ act by flipping $V$, i.e., $\tau\!\cdot\!(u,v)=(u,1-v)$. Then (A1) is false, i.e., $\mathbb{P}(Y\!=\!1\mid X\!=\!(u,v))=v$ is not $G$-invariant.
Now consider three representations.
\[
Z_{\text{full}}=X,\qquad
Z_{\text{good}}=V,\qquad
Z_{\text{bad}}=U\ \ (\text{$G$-invariant w.r.t.\ $\tau$}).
\]
For any proper scoring rule $\ell$, Bayes risks satisfy
\[
\mathcal{R}^\star(Z_{\text{full}})=0,
\qquad
\mathcal{R}^\star(Z_{\text{good}})=0,
\qquad
\mathcal{R}^\star(Z_{\text{bad}})=\tfrac{1}{2}.
\]
Indeed, given $Z_{\text{bad}}=U$, the posterior $\mathbb{P}(Y\!=\!1\mid U)=\tfrac12$, so one cannot beat random guess (for log-loss, $H(Y|U)=1$ vs. $H(Y|V)=0$). Here, enforcing invariance to flipping $V$ (by mapping both $(u,0)$ and $(u,1)$ to the same code) destroys label information and doubles the Bayes risk from $0$ to $1/2$.

Next, we consider a vision-flavored variant (``6 vs 9''). Let $G$ be $180^\circ$ rotations on images. If the label is the digit identity as written (``6'' vs ``9''), then $\mathbb{P}(Y\mid X)$ is not invariant to $180^\circ$ rotation: rotating a ``6'' yields a ``9''. Any representation forced to be invariant to $180^\circ$ rotation merges these two classes and raises Bayes risk to chance level. This illustrates that the correct invariance set must be task-true (and satisfy A1). Thus, PeL should target nuisances, not causal factors.

\section{Conclusion}\label{sec:concl}
In this work, we have framed Perception Learning (PeL) as a distinct optimization paradigm, equipping it with dedicated objectives, task-agnostic metrics, and a rigorous theoretical foundation that decouples sensory representation learning from downstream decision processes. By targeting perceptual properties through a minimal three-term loss—balancing invariance, contrastive discriminability, and diversity—we enable the creation of reusable, robust codes $Z$ that enhance modularity and transferability in AGI systems. The orthogonality theorem demonstrates that, under task-true invariances, PeL refinements preserve Bayes-optimal performance. This foundation paves the way for perception-first architectures that align with psychological principles of sensory adaptation, complementing self-supervised methods while addressing their evaluation limitations.

\end{document}